\documentclass[conference]{IEEEtran}
\IEEEoverridecommandlockouts
\usepackage{cite}
\usepackage{amsmath,amssymb,amsfonts}
\usepackage{algorithmic}
\usepackage{graphicx}
\usepackage{textcomp}
\usepackage{xcolor}
\usepackage{booktabs}
\usepackage{dsfont}
\usepackage{multirow}
\def\BibTeX{{\rm B\kern-.05em{\sc i\kern-.025em b}\kern-.08em
    T\kern-.1667em\lower.7ex\hbox{E}\kern-.125emX}}
\begin{document}

\title{From Deterministic to Probabilistic: A Novel Perspective on Domain Generalization for Medical Image Segmentation}

\author{Yuheng Xu \and Taiping Zhang }

\maketitle

\begin{abstract}
Traditional domain generalization methods often rely on domain alignment to reduce inter-domain distribution differences and learn domain-invariant representations. However, domain shifts are inherently difficult to eliminate, which limits model generalization. To address this, we propose an innovative framework that enhances data representation quality through probabilistic modeling and contrastive learning, reducing dependence on domain alignment and improving robustness under domain variations. Specifically, we combine deterministic features with uncertainty modeling to capture comprehensive feature distributions. Contrastive learning enforces distribution-level alignment by aligning the mean and covariance of feature distributions, enabling the model to dynamically adapt to domain variations and mitigate distribution shifts. Additionally, we design a frequency-domain-based structural enhancement strategy using discrete wavelet transforms to preserve critical structural details and reduce visual distortions caused by style variations. Experimental results demonstrate that the proposed framework significantly improves segmentation performance, providing a robust solution to domain generalization challenges in medical image segmentation.
\end{abstract}


\section{Introduction}
\label{sec:intro}
In modern medicine, medical image segmentation is a critical technology for accurately delineating tissues, organs, or lesions, providing essential support for disease diagnosis, treatment planning, and progression monitoring. Despite significant advancements in deep learning, its practical application in image segmentation faces considerable challenges. Among these, domain shift is particularly prominent, characterized by substantial distributional differences between training data (source domain) and testing data (target domain). These discrepancies are often caused by variations in imaging protocols, equipment types, operator techniques, and patient-specific characteristics, severely undermining the cross-domain robustness and adaptability of segmentation models.

\begin{figure}[t]
	\centering
	\includegraphics[width=1\linewidth]{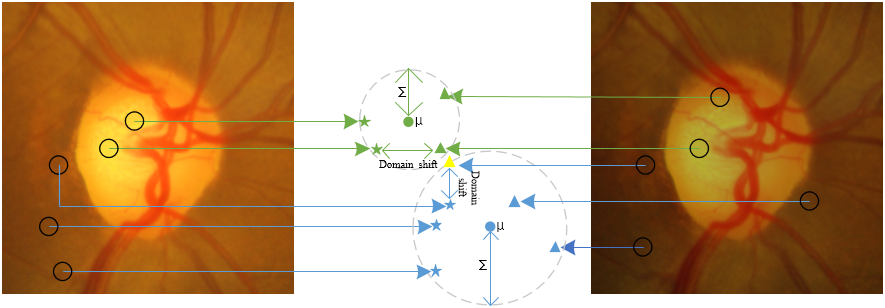} 
	\caption{Traditional domain alignment methods struggle to completely eliminate domain shifts, leading to sensitivity to feature distribution deviations in the model. In this figure, features of the same category exhibit significant distribution differences across different domains. Direct use of point estimation may result in inaccurate classifications. For example, the yellow sample in the middle is incorrectly classified into the green category because it is closer to the green prototype. However, by introducing probabilistic modeling and uncertainty estimation, the model can identify the differences in feature distributions. With the consideration of uncertainty, the yellow sample is more appropriately classified into the blue category, thereby improving the model’s adaptability and classification accuracy under domain shift conditions.}
	\label{fig1}
\end{figure}
Existing approaches to address domain shift often rely on domain alignment strategies \cite{muandet2013domain,li2018domain,li2018deep}, which minimize inter-domain feature distribution discrepancies to learn domain-invariant representations. However, these methods have inherent limitations (as shown in Fig. \ref{fig1}). First, domain shift is fundamentally difficult to eliminate, and attempts to explicitly bridge domain discrepancies are often impractical in real-world scenarios. Second, over-reliance on domain alignment can neglect critical features, particularly when source and target domains exhibit significant variability, making it challenging to capture fine-grained details or complex patterns. Moreover, these methods often overlook the importance of improving the quality of data representation, which is critical for ensuring model robustness and generalization under inevitable domain shift conditions.

To overcome these limitations, we propose an innovative solution that emphasizes improving data representation quality through the combined use of probabilistic modeling and contrastive learning, rather than relying heavily on domain alignment. Specifically, we construct a probabilistic representation learning framework that integrates deterministic features with uncertainty modeling in the latent space. By capturing the mean and variance of feature distributions, the framework quantifies uncertainty and enables dynamic adaptation to inter-domain variations. Combined with contrastive learning, the framework enforces distributional constraints across domains, allowing the model to learn more robust feature representations. This approach effectively mitigates cross-domain feature distribution shifts and significantly enhances predictive reliability, particularly in complex scenarios such as ambiguous boundaries, small lesions, or incomplete data. Compared to traditional methods, our approach redefines the solution to domain shift by enhancing data representation quality through probabilistic modeling and contrastive learning, eliminating the reliance on explicit domain alignment. This novel and efficient solution addresses the domain generalization challenge in medical image segmentation, offering both theoretical significance and practical application potential.

Furthermore, style transfer \cite{zhou2022generalizable}, as a data augmentation technique, has been widely applied in domain generalization to increase the diversity of training data and expand the coverage of the training domain, thereby enhancing model generalization. However, in some cases, certain styles (such as overly bright or dark transformations) can lead to blurred image structures, reducing edge sharpness and diminishing fine texture details, even altering the overall structural perception to some extent. For medical images, where semantic features are highly similar, such structural deviations may negatively impact the accuracy and robustness of segmentation models. To address these issues, we propose a structure enhancement strategy based on the frequency domain to compensate for structural detail distortions introduced by style transfer. Specifically, this method utilizes discrete wavelet transform to separate style features from high-frequency structural information in the image and reinforces high-frequency components in intermediate transformation results by incorporating high-frequency details from source domain images. This approach effectively preserves structural details by retaining critical structural information at both global and local scales, thereby enhancing edge and texture features. As a result, the model’s robustness is improved in cross-domain scenarios.

Extensive experiments have validated the effectiveness of the proposed module on two standard domain-generalized medical image segmentation benchmarks. Compared to previous methods on benchmark datasets, our framework achieves superior performance.  Our contributions can be summarized as follows: 
\begin{itemize}
    \item We propose a probabilistic representation learning framework that models both certainty and uncertainty within the latent space of images, prioritizing the enhancement of data representation quality rather than eliminating domain shift. By improving the robustness of feature representations, this approach enables the model to perform reliably under inevitable domain shift conditions.
    \item We developed a frequency-domain-based structural enhancement strategy to address the adverse effects of certain styles on the visual integrity of structural information during style transformations.
\end{itemize}

\section{Related Work}

\noindent\textbf{Domain Generalization} 
aims to train models capable of achieving strong generalization performance on unseen target domains by leveraging data from one or multiple source domains \cite{wang2022generalizing}. To this end, researchers have explored various techniques \cite{zhou2022domain}, such as data augmentation \cite{chen2023treasure,zhu2024structural,zhou2024mixstyle}, adversarial training \cite{qiao2020learning}, contrastive learning \cite{kim2021selfreg,xie2023sepico}, and meta-learning \cite{li2018learning}, which have led to significant advancements, particularly in the field of medical imaging. However, most existing methods rely heavily on domain alignment, aiming to reduce distributional differences across source domains to learn domain-invariant feature representations. These approaches assume that domain alignment can fully mitigate the impact of domain shifts. In practice, however, the complexity of inter-domain differences often renders domain shifts difficult to eliminate entirely. Residual domain shifts may result in biased feature representations, thereby limiting the model's generalization performance in target domains. Instead of over-relying on domain alignment, a more effective approach focuses on enhancing the quality of feature representations. By employing more robust feature modeling strategies, models can improve their robustness and adaptability, maintaining strong performance even under domain shift conditions.

\begin{figure*}[t]
	\centering
	\includegraphics[width=1\textwidth]{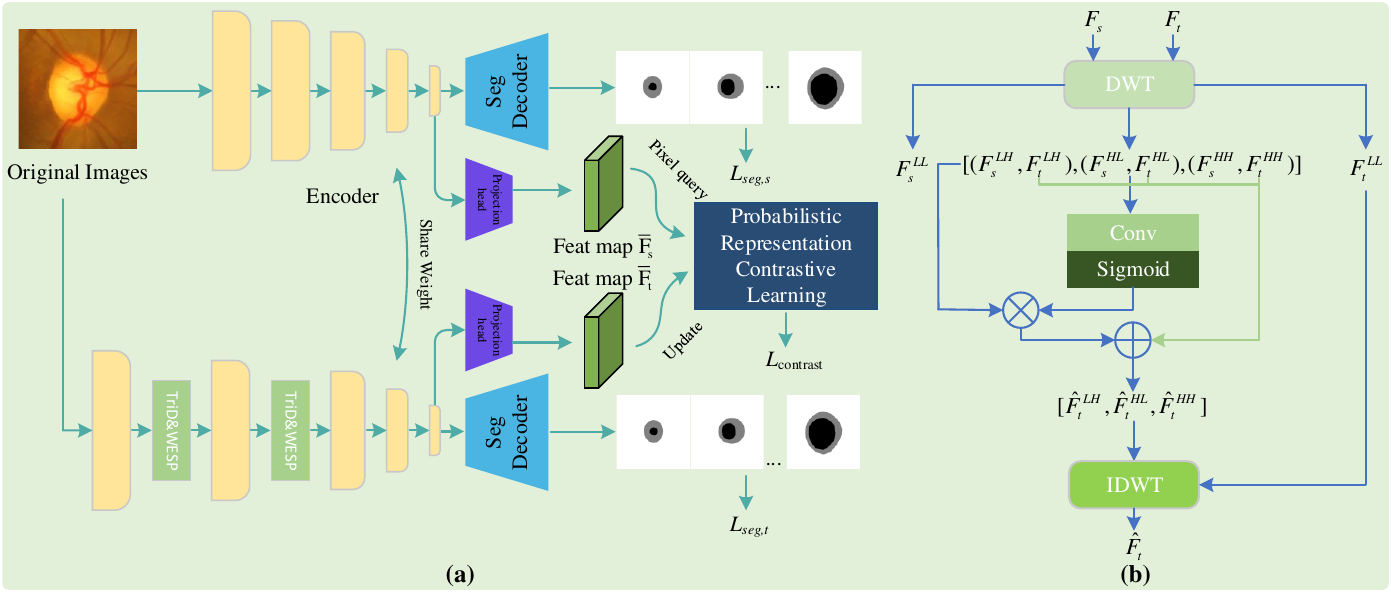} 
	\caption{The overall architecture of our method. (a)Probabilistic Representation Learning: We construct probabilistic representation learning by leveraging both the certainty and uncertainty in the latent space of images to explore the intrinsic relationships within the training data, mitigate domain shift, and fully realize the potential of the model. (b)Wavelet-Enhanced Structural Preservation: Using discrete wavelet transform (DWT) to separate style features from high-frequency structural information in the image, this method incorporates high-frequency components from source domain images into intermediate transformation results to enhance the expression of high-frequency information, thereby effectively preserving structural details.}
	\label{fig2}
\end{figure*}

\section{Method}
The overview of our framework is illustrated in the Fig. \ref{fig2}. Given a set of $D$ source domains ($\left\{\left(x_{i}^{d}, y_{i}^{d}\right)_{i=1}^{N_{d}}\right\}_{d=1}^{D}$), our objective is to endow the medical image segmentation models the capability to extract domain-invariant representations.
Here, $x_{i}^{d}$ is the $i^{t h}$ image from $d^{t h}$ source domain, $y_{i}^{d}$ is the segmentation label of $x_{i}^{d}$, and $N_{d}$ is the number of samples in $d^{t h}$ source domain.

\subsection{Probabilistic Representation Learning}

\subsubsection{Statistics Modeling}
Given the feature map $\bar{F}_{s} \in \mathbb{R}^{{\bar{C} } \times \bar{H} \times \bar{W}}$ and $\bar{F}_{t} \in \mathbb{R}^{{\bar{C} } \times \bar{H} \times \bar{W}}$ extracted from the source image and augmented image, we first partition these two features into $k^{t h}$ semantic classes based on the groundtruth labels $M_{i} \in \mathbb{R}^{ \bar{H} \times \bar{W}}$  to obtain $\Lambda^{k}_s$ and $\Lambda^{k}_t$ . 
Then, the local mean of the $k$-th class in the image is calculated to obtain $\mu_s^{\prime k}$. Afterwards, the class mean is calculated online across the entire source domain and augmented domain:
\begin{equation}\label{eq1}
	\mu_s^{\prime k}=\frac{1}{\left|\Lambda_s^{k}\right|} \sum_{p \in\left\{1,2, \cdots, H^{\prime} \times W^{\prime}\right\}} \mathds{1}_{\left[M_{i}=k\right]} \bar{F}_{s, p},
\end{equation}
where $|\cdot|$ is the cardinality of the set. To obtain global class prototypes, we choose to adopt an online approach across the entire source domain, iteratively aggregating and averaging statistics to construct the global class prototypes.
\begin{equation}
	\label{eq2}
	\mu_{s(j)}^{k}=\frac{n_{s(j-1)}^{k} \mu_{s(j-1)}^{k}+m_{s(j)}^{k} \mu^{\prime k}{}_{s(j)}}{n_{s(j-1)}^{k}+m_{s(j)}^{k}},
\end{equation}
where $m_{(j)}^{k}$ represents the total number of pixels belonging to the $k$-th class in the current $j$-th image and $n_{(j-1)}^{k}$ represents the total number of pixels belonging to the $k^{t h}$ class in the previous $j-1$ images. 
The class mean encapsulates the overall central tendency of features for each semantic class, providing a global representation of the class. However, relying solely on the mean vector is insufficient to capture the full variability within the feature space, particularly under domain shift conditions. Domain shifts often result in significant changes in the dispersion and relationships between feature dimensions in the target domain, which cannot be adequately captured by the mean alone. To address this limitation, we incorporate covariance modeling, which encodes the relationships and dependencies between feature dimensions. The covariance matrix complements the mean vector by providing a more comprehensive representation of feature distributions, enabling the model to dynamically adapt to variations in dispersion or orientation between the source and target domains. The covariance matrix also quantifies feature uncertainty, allowing the model to mitigate the effects of domain shift by down-weighting overly uncertain features and focusing on more robust patterns:
\begin{align}\label{eq3}
	\Sigma_{s(j)}^{k} = & \frac{n_{s(j-1)}^{k} m_{s(j)}^{k}(\mu_{s(j-1)}^{k}-\mu^{\prime k}_{s(j)})(\mu_{s(j-1)}^{k}-\mu^{\prime k }_{s(j)})^{\top}}{(n_{s(j-1)}^{k}+m_{s(j)}^{k})^{2}} \notag \\
	& + \frac{n_{s(j-1)}^{k} \Sigma_{s(j-1)}^{k} + m_{s(j)}^{k} \Sigma_{s(j)}^{\prime k}}{n_{s(j-1)}^{k}+m_{s(j)}^{k}},
\end{align}
where $\Sigma_{(j)}^{\prime k}$ is the covariance matrix between the $k^{t h}$ class in the $j^{t h}$ image. It is worth noting that the $\mu_{s(j)}^{k}$ and $\Sigma_{s(j)}^{k}$ are initialized to zero.
During training, we utilize feature maps $\bar{F}_{t}$ from synthetic images and dynamically update these statistical data using (\ref{eq2}) and (\ref{eq3}).

\subsubsection{Probabilistic Representation Contrastive Learning}
The contrastive loss is designed to ensure that features from the same class remain closely clustered while features from different classes are well-separated in the embedding space. By integrating probabilistic modeling, the contrastive loss is extended from point-to-point comparison to distribution-level alignment, incorporating both mean and covariance of feature distributions. This enables the model to account for inter-domain variability and uncertainty, ensuring that feature representations are not only class-consistent but also robust to domain shifts.
Specifically, we define the representation of any pixel in the source feature map as a pixel query $q \in \mathbb{R}^{\bar{C}}$, and we need to achieve lower loss values while simultaneously forming multiple positive pairs $\left(q, q_{m}^{+}\right) $ and negative pairs $\left(q, q_{n}^{k-}\right)$. Here, $q_{m}^{+}$ represents the $m^{th}$ positive example from the same category as the query $q$, while $q_{n}^{k-}$ represents the $n^{th}$ negative example from a different category $k$. We define a pixel  contrastive loss function for $q$ as follows:
\begin{equation}\label{eq4}
\ell_{q} = -\frac{1}{M} \sum\limits_{m=1}^{M} \log \frac{e^{q^{\top} q_{m}^{+} / \tau}}{e^{q^{\top} q_{m}^{+} / \tau} + \sum\limits_{k \in \mathcal{K}^{-}} \frac{1}{N} \sum\limits_{n=1}^{N} e^{q^{\top} q_{n}^{k-} / \tau}},
\end{equation}
where $M$ and $N$ represent the number of positive and negative pairs, respectively, and $\mathcal{K}^{-}$ denotes the set containing all classes different from $q$. When $M$ and $N$ are large, the computation can quickly exhaust GPU memory and reduce efficiency. Therefore, we have derived a special form of  contrastive loss to address this issue. The detailed derivation process can be found in the appendix.
\begin{equation}\label{eq5}
	\ell_{q} = -\log \frac{e^{\frac{q^{\top} \mu^{+}}{\tau} + \frac{q^{\top} \Sigma^{+} q}{2 \tau^{2}}}}{e^{\frac{q^{\top} \mu^{+}}{\tau} + \frac{q^{\top} \Sigma^{+} q}{2 \tau^{2}}} + \sum\limits_{k \in \mathcal{K}^{-}} e^{\frac{q^{\top} \mu^{k-}}{\tau} + \frac{q^{\top} \Sigma^{k-} q}{2 \tau^{2}}} },
\end{equation}
where $\mu^{+}$ and $\Sigma^{+}$ are the the mean and covariance of the positive semantic distribution of $q$, respectively. $\mu^{k-}$ and $\Sigma^{k-}$ are the mean and covariance of the negative distribution of $k^{t h}$, respectively. Overall, the semantic-guided pixel-wise contrastive loss is defined as follows:
\begin{equation}\label{eq6}
	\mathcal{L}^{d}_{ {contrast }}=\frac{1}{|\Psi|} \sum_{q \in \bar{F}^{d}_{s}} \ell_{q},
\end{equation}
where ${|\Psi|}$ is the total number of pixels in $\bar{F}^{d}_{s}$.

\subsection{Wavelet-Enhanced Structural Preservation}
We first use the style transformation method proposed by TriD \cite{chen2023treasure}, which modifies the statistical data of image features to generate augmented images features and thereby expand the coverage of the training domain. To compensate for the potential loss of structural details during style transformation, we further introduce the Discrete Wavelet Transform (DWT) to separate style features from high-frequency structural information in the images. We then use high-frequency information from the source domain images to enhance the high-frequency components of the transformed results, optimizing the preservation of structural details. Specifically, we input the source domain image features $F_{s} \in \mathbb{R}^{c \times h \times w}$ and the style-transformed features $F_{t} \in \mathbb{R}^{c \times h \times w}$ into the DWT, decomposing them into four wavelet sub-bands:
\begin{equation}\label{eq7}
\begin{aligned}
\left\{F_{s}^{LL}, F_{s}^{LH}, F_{s}^{HL}, F_{s}^{HH}\right\} &= DWT\left(F_{s}\right), \\
\left\{F_{t}^{LL}, F_{t}^{LH}, F_{t}^{HL}, F_{t}^{HH}\right\} &= DWT\left(F_{t}\right).
\end{aligned}
\end{equation}

Since the $F_{t}^{LL}$ sub-band primarily contains style information while the other three high-frequency sub-bands capture structural information, we use the corresponding high-frequency sub-bands of $F_{s}$ to enhance the three high-frequency sub-bands of $F_{t}$. This enhancement process is defined as follows:
\begin{equation}\label{eq8}
\hat{F}_{t}^{LH} = F_{t}^{LH} + F_{s}^{LH} \cdot \operatorname{Sigmoid}\left(\operatorname{Conv}\left(\left[F_{t}^{LH}, F_{s}^{LH}\right]\right)\right), 
\end{equation}
where $\hat{F}{t}^{HL}$ and $\hat{F}{t}^{HH}$ are processed similarly, applied respectively to $F_{t}^{HL}$ and $F_{t}^{HH}$.
After obtaining the enhanced high-frequency sub-bands, we reconstruct the image features through the Inverse Discrete Wavelet Transform (IDWT), as follows:
\begin{equation}\label{eq9}
\hat{F}_{t}=I D W T\left(F_{t}{ }^{L L}, \hat{F}_{t}^{L H}, \hat{F}_{t}^{H L}, \hat{F}_{t}^{H H}\right).
\end{equation}

\noindent\textbf{Discussions.} In this study, we opted for Discrete Wavelet Transform (DWT) over Fourier Transform (FT) to separate and enhance the high-frequency structural information in images. The main advantage of DWT lies in its multi-scale decomposition capability and time-frequency localization properties, which enable effective separation of image details and style features. Specifically, DWT’s multi-scale sub-band decomposition allows different frequency components of an image to be precisely localized in the spatial domain, enabling the enhancement of high-frequency structural information while preserving local spatial details. In contrast, Fourier Transform operates solely in the frequency domain, lacking spatial localization capabilities and therefore struggling to capture local detail features within images. Furthermore, DWT’s edge-preserving characteristics and noise-resilience are particularly valuable in cross-domain image style transfer tasks, as they allow structural details to be reinforced while suppressing noise, thus improving the quality and stability of generated images. In summary, the time-frequency localization, multi-scale decomposition, and noise-resilience of DWT make it more suitable than Fourier Transform for the task of separating and enhancing structural information in this study.

\subsection{Loss Function}
We employ both the Dice loss \cite{milletari2016v} and the unified cross-entropy (CE) loss \cite{murphy2012machine}  for segmentation losses. The overall loss function is described as follows:
\begin{align}\label{eq10}
\mathcal{L}_{{total}}=& \frac{1}{D} \sum_{d=1}^{D}(\mathcal{L}_{{seg,s}}^{d}+\mathcal{L}_{{seg,t}}^{d}+\mathcal{L}_{contrast}^{d}).
\end{align}

\begin{table*}[htb]
\centering
\caption{Dice coefficient (\%) and Average Surface Distance (voxel) produced by different methods on the \textbf{Fundus} dataset. The best results are bold-faced.}
\resizebox{1.0\textwidth}{!}{%
\begin{tabular}{@{}c|cccc|c|cccc|c@{}}
\toprule
\multirow{2}{*}{Task} & \multicolumn{5}{c|}{Optic Cup/Disc Segmentation} & \multicolumn{5}{c@{}}{Optic Cup/Disc Segmentation} \\ \cmidrule(lr){2-6} \cmidrule(lr){7-11}
& Domain 1 & Domain 2 & Domain 3 & Domain 4 & Avg & Domain 1 & Domain 2 & Domain 3 & Domain 4 & Avg \\ \midrule
RAM \cite{zhou2022generalizable} & 83.08/95.59 & 76.98/87.30 & 84.91/94.86 & 84.45/93.65 & 87.60 & 17.92/7.90 & 16.99/18.84 & 10.51/7.38 & 9.18/7.49 & 12.03 \\
CDDSA \cite{gu2023cddsa} & 85.78/\textbf{96.36} & 78.38/91.81 & 86.26/95.10 & 83.18/92.59 & 88.68 & 14.84/\textbf{6.33} & 14.48/11.84 & 9.73/7.04 & 9.68/8.13 & 10.26 \\
LRSR \cite{chen2024learning} & 84.55/95.45 & 77.00/87.94 & 86.10/95.26 & 84.77/93.68 & 88.09 & 16.34/8.07 & 16.12/16.99 & 10.07/6.93 & 9.60/7.81 & 11.49 \\
CSU \cite{zhang2024domain} & 79.53/92.77 & 79.85/90.13 & 85.51/93.27 & 85.98/94.12 & 87.65 & 20.79/10.21 & 15.46/12.44 & 10.88/10.35 & 8.03/7.40 & 11.95 \\
DeTTA \cite{wen2024denoising}&81.14/93.87 & \textbf{80.01}/90.21 & 86.29/93.51 & 86.18/94.59 & 88.23 & 19.94/8.91 & 13.98/11.86 & 9.96/10.06 & 7.97/6.81 & 11.19 \\
WT-PSE\cite{chen2024learning} & 86.71/96.23 & 79.64/\textbf{92.67} & \textbf{87.90}/93.69 & 83.75/93.04 & 89.20 & 14.10/6.76 & 13.46/\textbf{10.50} & 8.97/8.84 & 9.49/8.85 & 10.12 \\ \midrule
Ours & \textbf{86.87}/96.31 & 79.73/91.57 & 87.81/\textbf{95.27} & \textbf{88.01}/\textbf{94.86} & \textbf{90.05} & \textbf{14.02}/6.49 & \textbf{13.26}/11.03 & \textbf{8.93}/\textbf{6.84} & \textbf{6.98}/\textbf{5.86} & \textbf{9.18} \\
\bottomrule
\end{tabular}
}
\label{tab1}
\end{table*}

\begin{table*}[htb]
\centering
\caption{Dice coefficient (\%) and Average Surface Distance (voxel) produced by different methods on the \textbf{Prostate} dataset. The best results are bold-faced.}
\begin{tabular}{@{}c|cccccc|c@{}}
\toprule
Task     & \multicolumn{6}{c|}{Prostate Segmentation(Dice coefficient / Average Surface Distance)}   & \multirow{2}{*}{\centering Avg} \\ \cmidrule(r){1-7}
Domain   & Domain 1 & Domain 2 & Domain 3 & Domain 4 & Domain 5 & Domain 6 & \multicolumn{1}{c}{}                     \\ \midrule
RAM\cite{zhou2022generalizable} & 87.19/2.07    & 88.92/1.59    & 86.13/2.04    & 88.46/1.56    & 87.26/1.82    & 87.67/1.33    & 87.61/1.74                     \\
CDDSA \cite{gu2023cddsa}  & 88.65/1.84    & 88.91/1.76    & 86.58/2.13    & 88.35/1.48    & 86.71/1.98    & 88.24/1.16    & 87.91/1.73                                    \\
LRSR \cite{chen2024learning} & 86.84/2.21    & 88.53/1.84    & 86.22/1.97    & 88.94/1.68    & 86.21/2.08    & 87.58/1.63    & 87.39/1.90                                    \\
CSU \cite{zhang2024domain}   & 86.14/2.08    & 88.62/1.33    & 86.31/2.48    & 88.16/1.97    & 87.05/2.04    & 86.74/2.16    & 87.17/2.01                                 \\
DeTTA \cite{wen2024denoising}    & 88.23/1.47    & 88.14/1.51    & 87.82/2.13    & 88.91/1.38    & 87.53/1.86    & 88.79/1.37    & 88.24/1.62                       \\
WT-PSE\cite{chen2024learning}  & 89.65/1.23    & 89.34/1.19    & 86.57/2.21    & 89.74/1.01    & 88.25/1.41    & 89.97/1.08    & 88.92/1.41                                   \\ \midrule
Ours     & \textbf{91.63}/\textbf{0.63}    & \textbf{90.84}/\textbf{0.68}    & \textbf{88.18}/\textbf{1.93}    & \textbf{90.18}/\textbf{0.72}    & \textbf{88.28}/\textbf{1.28}    & \textbf{90.55}/\textbf{0.59}    & \textbf{89.94}/\textbf{0.97}                                    \\
\bottomrule
\end{tabular}   
\label{tab2}
\end{table*}

\begin{figure}[b]
	\centering
	\includegraphics[width=1\linewidth]{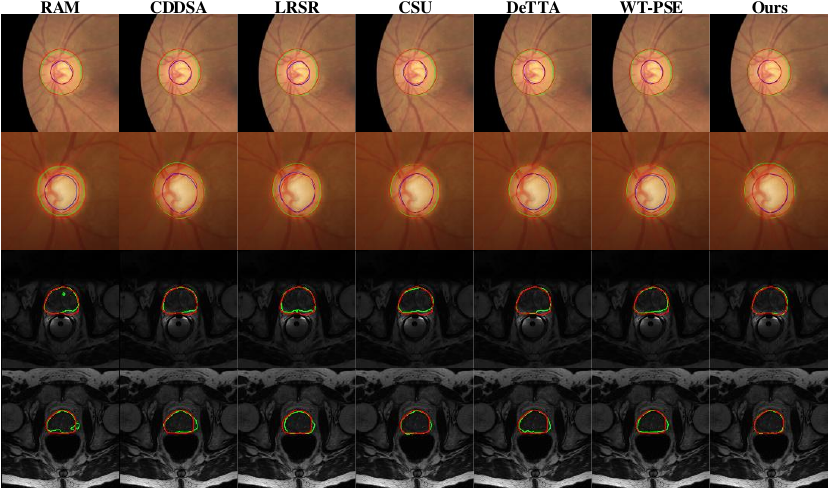} 
	\caption{Visual comparison for \textbf{Fundus} and \textbf{Prostate} segmentation task. The red contours indicate the boundaries of ground truths while the green and blue contours are predictions.}
	\label{fig3}
\end{figure}

\begin{table*}[h]
	\centering
	\caption{Ablation Study of key components in our method on the Fundus dataset (\%). }
	\label{tab3}
    \begin{tabular}{@{}ccc|cccc|c@{}}
\toprule
\multicolumn{3}{c|}{Task} & \multicolumn{4}{c|}{Optic Cup/Disc Segmentation}      & \multirow{2}{*}{Avg} \\ \cmidrule(r){1-7}
PRL(no covariance)    & PRL(+ covariance)    & WESP    & Domain 1    & Domain 2    & Domain 3    & Domain 4    &                      \\ \midrule
-          & -         & -         & 78.82/92.75 & 72.79/89.62 & 83.33/92.69 & 83.04/91.97 & 85.63  \\
\checkmark & -         & -         & 80.38/93.54 & 74.91/89.84 & 85.08/93.31 & 84.28/92.15 & 86.69  \\
\checkmark &\checkmark & -         & 84.16/95.02 & 77.29/90.87 & 86.88/94.79 & 87.83/93.37 & 88.78  \\
\checkmark &\checkmark &\checkmark & 86.87/96.31 & 79.73/91.57 & 87.81/95.27 & 88.01/94.86 & 90.05  \\ \bottomrule
\end{tabular}
\end{table*}

\section{Experiments}
\subsection{Datasets}
Two publicly available medical image segmentation datasets are employed in our experiments, including the \textbf{Fundus} dataset \cite{wang2020dofe} and the \textbf{Prostate} dataset \cite{liu2020shape}. The \textbf{Fundus} dataset comprises retinal fundus images sourced from different scanners at four different medical centers, serving the purpose of optic cup and disc segmentation tasks. Each domain within this dataset was further divided into training and testing sets. Following the pre-processing protocol in \cite{wang2020dofe}, each image was center cropped the disc regions using $800 \times 800$ bounding boxes. Subsequently, random resizing and cropping were performed to obtain a $256 \times 256$ region from each cropped image, which served as input for our network.The \textbf{Prostate} dataset contains \textbf{T2}-weighted MRI prostate images gathered from six distinct data sources, primarily for prostate segmentation. The $3D$ prostate region were cropped and $2D$ slices in the axial plane have been uniformly resized to $384 \times 384$. During model training, we input $2D$ slices of the prostate images into our model. Data normalization has been individually applied to both datasets, resulting in intensity values scaled to the range [-1, 1].


\subsection{Implementation Details}
We adopt a U-shaped segmentation network \cite{hu2022domain} with a ResNet-34 \cite{he2016deep} backbone, conducting experiments on an NVIDIA 3090 GPU. Using a batch size of 16, we train the model for 200 epochs with an SGD optimizer and a momentum of 0.99 for both tasks. Initial learning rates are set to 0.001 for the prostate dataset and 0.004 for the OD/OC dataset, with a polynomial decay schedule to stabilize training. We follow the standard practice in domain generalization research by adopting the leave-one-domain-out strategy, where the model is trained on $K$ source domains and tested on the remaining target domain (total $K+1$ domains). For evaluation, we use the Dice coefficient (Dice) and Average Surface Distance (ASD), following conventions in medical image segmentation. To ensure consistency, each experiment was repeated three times, with the average performance reported.

\subsection{Comparison with Other methods}
Tables \ref{tab1} and \ref{tab2} compare the performance of our proposed method with state-of-the-art approaches on the Fundus and Prostate datasets, while Fig. \ref{fig3} visually highlights its advantages in accurate boundary segmentation.

For the Fundus dataset (Table \ref{tab1}), our method achieves an average Dice coefficient of 90.05\% and an average surface distance (ASD) of 9.18 voxels, outperforming the current best-performing method, WT-PSE, by 0.85\% in Dice and consistently maintaining lower ASD. Additionally, Fig \ref{fig2} demonstrates the method’s ability to handle complex boundary regions and accurately capture critical structures, emphasizing its strength in boundary modeling.

On the Prostate dataset (Table \ref{tab2}), our method attains an average Dice coefficient of 89.94\% and the lowest ASD of 0.97 voxels, significantly surpassing other methods. Compared to WT-PSE, it improves Dice by 1.02\% and reduces ASD by 0.44 voxels. These results, combined with the visual evidence from Fig \ref{fig2}, underline the robustness of our approach in cross-domain scenarios and its precise delineation of structural details.

Overall, the quantitative and qualitative results strongly validate the effectiveness of our method in achieving high-precision segmentation and superior boundary refinement, consistently outperforming existing state-of-the-art methods across diverse datasets and tasks.

\subsection{Ablation Study}
To evaluate the effectiveness of the key components in our proposed method, we conducted ablation experiments on the Fundus dataset for the optic cup and disc segmentation task. The results in Table~\ref{tab3} demonstrate the contribution of each component. The baseline model uses TriD for data augmentation without incorporating probabilistic representation learning (PRL) or wavelet-enhanced structural preservation (WESP), achieving an average Dice score of 85.63

Introducing PRL without covariance modeling (mean-based modeling only) improves the Dice score to 86.69\%, indicating that mean-based representations enhance feature separability and generalization. However, this approach is limited in capturing uncertainty and distribution diversity. Incorporating covariance modeling into PRL further boosts the Dice score to 88.78\%, demonstrating its ability to capture complex feature relationships and adapt to domain-specific variations by providing a comprehensive representation of feature distributions. Combining PRL with contrastive learning further aligns the mean and covariance across domains, improving domain invariance and robustness. Finally, adding WESP achieves the best performance, with an average Dice score of 90.05\%. WESP effectively preserves critical structural information during style transfer, mitigating visual distortions caused by extreme transformations. This ensures reliable feature extraction and significantly enhances segmentation performance.

\section{Conclusion}
In this paper, we propose a novel approach to address domain generalization challenges in medical image segmentation by integrating probabilistic modeling and frequency-domain structural enhancement. The proposed framework effectively enhances data representation quality, improves model robustness and generalization under domain shifts, and mitigates the adverse effects of style diversity. Experimental results demonstrate the method’s superiority and its potential for tackling real-world medical imaging challenges. Future work could explore the scalability of this approach to other medical imaging tasks and investigate further optimization of uncertainty modeling to improve performance in highly diverse scenarios.

\bibliographystyle{IEEEbib}
\bibliography{icme2025_template_anonymized}

\end{document}